\documentclass[11pt,a4paper]{article}
\usepackage{authblk}
\usepackage[hyperref]{emnlp-ijcnlp-2019}
\usepackage{times}
\usepackage{latexsym}
\usepackage{amsfonts}
\usepackage{url}
\usepackage{booktabs}
\usepackage{multirow}
\usepackage{algorithmic}
\usepackage{algorithm}
\usepackage{subcaption}
\usepackage{graphicx}
\usepackage{xspace}
\usepackage{tikz}
\usepackage{amsfonts, float, xcolor}
\usepackage{amsmath,amssymb}
\usepackage{systeme,mathtools,enumerate}
\usetikzlibrary{positioning,arrows.meta,quotes}
\usetikzlibrary{shapes,snakes}
\usetikzlibrary{bayesnet}
\tikzset{>=latex}
\tikzstyle{plate caption} = [caption, node distance=0, inner sep=0pt, below left=5pt and 0pt of #1.south]

\graphicspath{ {images/} }
\aclfinalcopy 

\usepackage[normalem]{ulem}

\newcommand{\name}{LaSyn\xspace}

\newcommand{\xy}[1]{\textcolor{blue}{$_{XY}$[#1]}}

\newcommand{\rev}[1]{\textcolor{blue}{{#1}}}
\newcommand{\eat}[1]{}

\title{Latent Part-of-Speech Sequences for Neural Machine Translation}

\author[1]{Xuewen Yang}
\author[1]{Yingru Liu}
\author[2]{Dongliang Xie}
\author[1]{Xin Wang}
\author[1,3]{Niranjan Balasubramanian}
\affil[1]{Stony Brook University}
\affil[2]{Beijing University of Posts and Telecommunications}
\affil[1]{\tt \{xuewen.yang,yingru.liu,x.wang\}@stonybrook.edu}
\affil[2]{\tt xiedl@bupt.edu.cn}
\affil[3]{\tt niranjan@cs.stonybrook.edu}

\date{}

\begin{document}
\maketitle

\begin{abstract}
Learning target side syntactic structure has been shown to improve Neural Machine Translation (NMT). However, incorporating syntax through latent variables introduces additional complexity in inference, as the models need to marginalize over the latent syntactic structures. To avoid this, models often resort to greedy search which only allows them to explore a limited portion of the latent space.

In this work, we introduce a new latent variable model, \name, that captures the co-dependence between syntax and semantics, while allowing for effective and efficient inference over the latent space. \name decouples direct dependence between successive latent variables, which allows its decoder to exhaustively search through the latent syntactic choices, while keeping decoding speed proportional to the size of the latent variable vocabulary.
We implement \name by modifying a transformer-based NMT system and design a neural expectation maximization algorithm that we regularize with part-of-speech information as the latent sequences. 
Evaluations on four different MT tasks show that incorporating target side syntax with \name improves both translation quality, and also provides an opportunity to improve diversity. 

\end{abstract}

\section{Introduction}

Syntactic information has been shown to improve the translation quality in NMT models. On the source side, syntax can be incorporated in multiple ways --- either directly during encoding~\cite{Chen18Syntax, SennrichLinguistic16, eriguchi2016tree}, or indirectly via multi-task learning to produce syntax informed representations~\cite{eriguchi2017learning, Baniata18, Niehues17, Zaremoodi18}. On the target side, however, incorporating the syntax is more challenging due to the additional complexity in inference when decoding over latent states. To avoid this, existing methods resort to approximate inference over the latent states using a two-step decoding process~\cite{gu2018top, wang2018tree, wu2017sequence, aharoni2017}. Typically, the first stage decoder produces a beam of latent states, which serve as conditions to feed into the second stage decoder to obtain the target words. Thus, training and inference in these models can only explore a limited sub-space of the latent states.  

In this work, we introduce \name, a new target side syntax model that allows exhaustive exploration of the latent states to ensure a better translation quality. Similar to prior work, \name approximates the co-dependence between syntax and semantics of the target sentences by modeling the joint conditional probability of the target words and the syntactic information at each position. However, unlike prior work, \name eliminates the sequential dependence between the latent variables and simply infers the syntactic information at a given position based on the source text and the partial translation context. This allows \name to search over a much larger set of latent state sequences. In terms of time complexity, unlike typical latent sequential models, \name only introduces an additional term that is linear in the size of latent variable vocabulary and the length of the sentence.


We implement \name by modifying a transformer-based encoder-decoder model. The implementation uses a hybrid decoder that predicts two posterior distributions: the probability of syntactic choices at each position $P(\mathbf{z}_n \vert \mathbf{x}, \mathbf{y}_{<n})$, and the probability of the word choices at each position conditioned on each of the possible values for the latent states $P(\mathbf{y}_n \vert \mathbf{z}_n, \mathbf{x}, \mathbf{y}_{<n})$. 
The model cannot be trained by directly optimizing the data log-likelihood because of its non-convex property. We devise a neural expectation maximization (NEM) algorithm, 
whose E-step computes the posterior distribution of latent states under current model parameters, and M-step updates model parameters using gradients from back-propagation. Given some supervision signal for the latent variables, we can modify this EM algorithm to obtain a regularized training procedure. We use parts-of-speech (POS) tag sequences, automatically generated by an existing tagger, as the source of supervision. 

Because the decoder is exposed to more latent states during training, it is more likely to generate diverse translation candidates. To obtain diverse sequences, we can decode the most likely translations for different POS tag sequences. This is a more explicit and effective way of performing diverse translation than other methods based on diverse or re-ranking beam search~\cite{Vijayakumar18, LiJ16mutual}, or coarse codes planning~\cite{Shu18}. 

We evaluate \name on four translation tasks. Evaluations show that \name outperforms models that only use partial exploration of the latent states for incorporating target side syntax. 
A diversity based evaluation also shows that when using different POS tag sequences during inference, \name produces more diverse and meaningful translations compared to existing models. 
\section{A Latent Syntax Model for Decoding}

In a standard sequence-to-sequence model, the decoder directly predicts the target sequence $\mathbf{y}$ conditioned on the source input $\mathbf{x}$. The translation probability $P(\mathbf{y}\vert \mathbf{x})$ is modeled directly using the probability of each target word $\mathbf{y}_n$ at time step $n$ conditioned on the source sequence $\mathbf{x}$ and the current partial target sequence $\mathbf{y}_{<n}$ as follows:
\begin{equation}
P(\mathbf{y} \vert \mathbf{x}; \boldsymbol{\theta}) = \prod_{n=1}^{N} P(\mathbf{y}_n\vert \mathbf{x}, \mathbf{y}_{<n};\boldsymbol{\theta})
\end{equation}
where, $\boldsymbol{\theta}$ denotes the parameters of both the encoder and the decoder.
 
In this work, we model syntactic information of target tokens using an additional sequence of variables, which captures the syntactic choices\footnote{The variables can be used to model any linguistic information that can be expressed as choices for each word position (e.g., morphological choices).} at each time step. There are multiple ways of incorporating this additional information in a sequence-to-sequence model.

\begin{figure}
\centering
\begin{subfigure}[b]{0.35\textwidth}
   \includegraphics[width=1\linewidth]{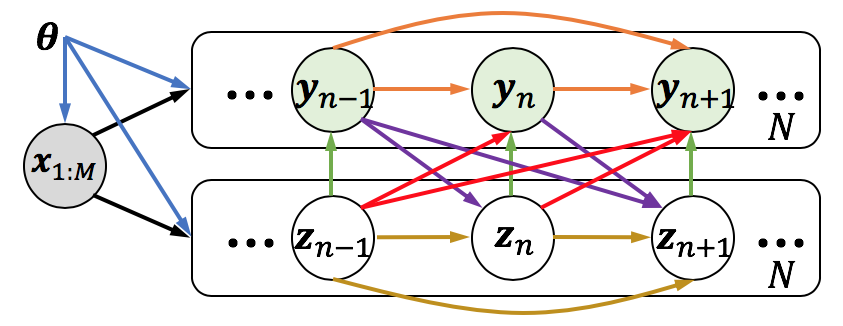}
   \caption{Full co-dependence model.}
   \label{fig:Bayes1} 
\end{subfigure}

\begin{subfigure}[b]{0.35\textwidth}
   \includegraphics[width=1\linewidth]{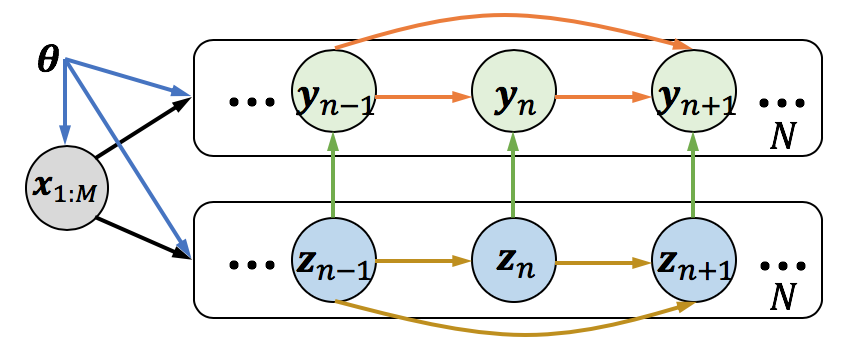}
   \caption{Two-step decoding model.}
   \label{fig:Bayes2}
\end{subfigure}

\begin{subfigure}[b]{0.35\textwidth}
   \includegraphics[width=1\linewidth]{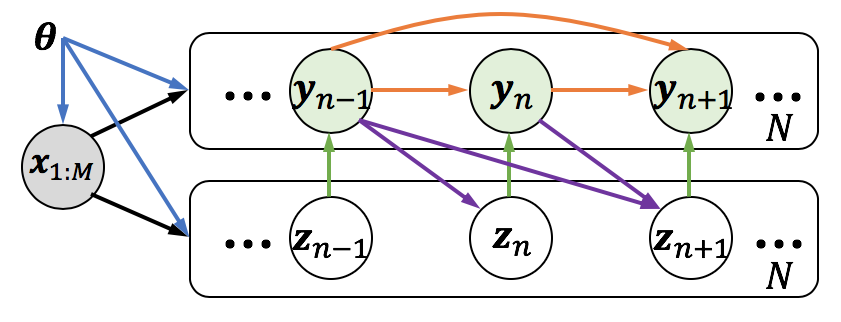}
   \caption{\name: Our Latent syntax model}
   \label{fig:Bayes3}
\end{subfigure}

\caption[]{Target-side Syntax Models: (a) An ideal solution that captures full co-dependence between syntax and semantics. (b) A widely-used two-step decoding model~\cite{wang2018tree, wu2017sequence, aharoni2017}. 
(c) \name, our latent syntax model that uses non-sequential latent variables for exhaustive search of latent states.}
\end{figure}
An ideal solution should capture the co-dependence\eat{\xy{or inter-dependence???} co-dependence is more accurate i think.} between syntax and semantics. In a sequential translation process, the word choices at each time step depend on both the semantics and the syntax of the words generated at the previous time steps. The same dependence also holds for the syntactic choices at each time step.
Figure~\ref{fig:Bayes1} shows a graphical model that captures this {\em full} co-dependence between the syntactic variable sequence $\mathbf{z}_1, \dots, \mathbf{z}_N$ and the output word sequence $\mathbf{y}_1, \dots, \mathbf{y}_N$. 
Such a model can be implemented using two decoders, one to decode syntax and the other to decode output words. The main difficulty, however, is that inference in this model is intractable since it involves marginalizing over the latent $\mathbf{z}$ sequences.

To keep inference tractable, existing approaches treat syntactic choices $\mathbf{z}$ as observed sequential variables~\cite{gu2018top, wang2018tree, wu2017sequence, aharoni2017}, as shown in Figure~\ref{fig:Bayes2}. 
These models use a two-stage decoding process, where for each time step they first produce most likely latent state $\mathbf{z}_n$ and then use this as input to a second stage that decodes words.
However, this process is unsatisfactory in two respects. First, the inference of syntactic choices is still approximate as it does not explore the full space of $\mathbf{z}$.
Second, these models are not well-suited for controllable or diverse translations. Using such a model to decode from an arbitrary $\mathbf{z}$ sequence is a divergence from its training, where it only learns to decode from a limited space of $\mathbf{z}$ sequences.

\subsection{Model Description}
Our goal is to design a model that allows for exhaustive search over syntactic choices.
We introduce \name, a new latent model shown in Figure~\ref{fig:Bayes3}. The syntactic choices are modeled as {\em true latent variables} i.e., unobserved variables. Compared to the ideal model in Figure~\ref{fig:Bayes1}, \name includes two simplifications for tractability: (i) The dependence between successive syntactic choices is modeled indirectly, via the word choices made in the previous time steps. (ii) The word choice at each position depends only on the syntactic choice at the current position and the previous predicted words. Dependence on previous syntactic choices is modeled indirectly.

Under this model, the joint conditional probability of the target word $\mathbf{y}_n$ together with its corresponding latent syntactic choice $\mathbf{z}_n$\footnote{Note that $\mathbf{z}_n \in V_{\mathbf{z}}$, where $V_{\mathbf{z}}$ is the vocabulary of latent syntax for the target, which differs from language to language. 
} is given by:

\begin{equation}
\begin{aligned}
P(\mathbf{y}_n, \mathbf{z}_n \vert \mathbf{x}, \mathbf{y}_{<n}) &= P(\mathbf{y}_{n} \vert \mathbf{z}_n, \mathbf{x}, \mathbf{y}_{<n}) \\
& \times P(\mathbf{z}_n\vert \mathbf{x}, \mathbf{y}_{<n})
\end{aligned}
\label{eq:joint}
\end{equation}

\begin{figure}
\centering
\includegraphics[width=0.5\textwidth]{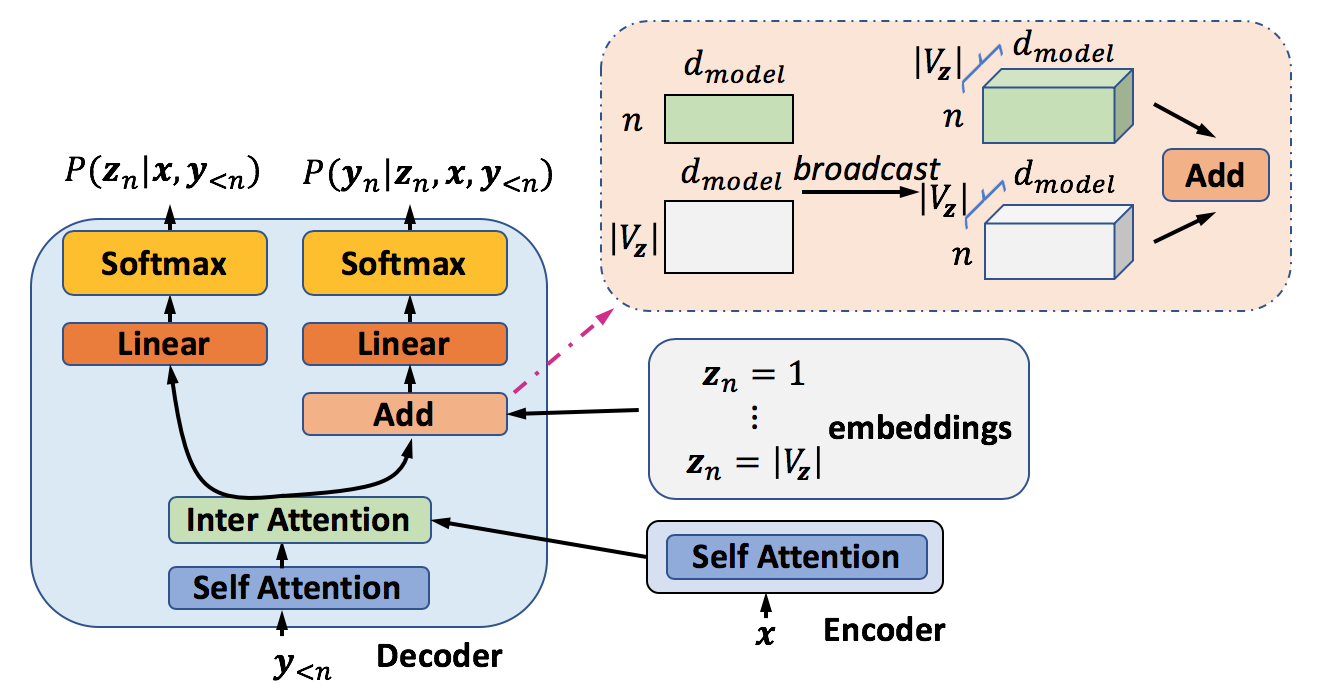}
\caption{The architecture of \name.}
\label{fig:architecture}
\end{figure}

We implement \name by modifying the Transformer-based encoder-decoder architecture~\cite{Vaswani17}. As shown in Figure~\ref{fig:architecture}, \name consists of a shared encoder for encoding source sentence $\mathbf{x}$ and a hybrid decoder that manages the decoding of the latent sequence $\mathbf{z}$ (left branch) and the target sentence $\mathbf{y}$ (right branch) separately. 

The encoder consists of the standard self-attention layer, which generates representations of each token in the source sentence $\mathbf{x}$.
The hybrid decoder consists of a self-attention layer that encodes the output generated thus far (i.e., the partial translation), followed by a inter-attention layer which computes the attention across the encoder and decoder representations. 

The decoder's left branch predicts the latent variable distribution $P(\mathbf{z}_n\vert \mathbf{x}, \mathbf{y}_{<n})$ by applying a simple linear transformation and softmax on the inter-attention output, which contains information about the encoded input $\mathbf{x}$ and the partial translation $\mathbf{y}_{<n}$. 

The right branch predicts the target word distribution $P(\mathbf{y}_{n} \vert \mathbf{z}_n, \mathbf{x}, \mathbf{y}_{<n})$ using the inter-attention output and the embeddings of all the available choices for $\mathbf{z}_n$. The choices for $\mathbf{z}_n$ are represented as embeddings that can be learned during training. We then combine the inter-attention output and the latent choice embeddings through an \texttt{Add} operation, which is a simple composition function that captures all combinations of additive interactions between the two. The dimension of the inter-attention is $n\times d_{model}$ and that of the latent embeddings is $\vert V_{\mathbf{z}} \vert \times d_{model}$, where $\vert V_{\mathbf{z}} \vert$ is the total number of choices for $\mathbf{z}_n$ or the size of the latent variable vocabulary. We broadcast them to the same dimension $n \times \vert V_{\mathbf{z}} \vert \times d_{model}$ and then simply add them together point-wise as shown in Figure~\ref{fig:architecture}. 
This is then fed to a linear transform and softmax over the target word vocabulary.

\subsection{Inference with Exhaustive Search for Latent States}

When using additional variables to model target side syntax, exact inference requires marginalizing over these additional variables.
\begin{equation}
   P(\mathbf{y} \vert \mathbf{x}) = \sum_{\mathbf{z}\in F(\mathbf{z})} P(\mathbf{y} \vert \mathbf{z},\mathbf{x})P(\mathbf{z} \vert \mathbf{x}) 
\end{equation}

To avoid this exponential complexity, prior works use a two-step decoding process with models similar to the one shown in Figure~\ref{fig:Bayes2}. They use greedy or beam search to explore a subset $B(\mathbf{z})$ of the latent space to compute the posterior distribution as follows:
\begin{equation}
   P(\mathbf{y} \vert \mathbf{x}) \simeq \sum_{\mathbf{z}\in B(\mathbf{z})} P(\mathbf{y} \vert \mathbf{z},\mathbf{x})P(\mathbf{z} \vert \mathbf{x}) 
\end{equation}

\eat{When modeling the target side syntactic structure, to our best knowledge, most of the existing methods~\cite{wang2018tree, wu2017sequence, aharoni2017} used a Bayesian architecture similar to Figure~\ref{fig:Bayes2}. Because of the exponential search space of sequence $\mathbf{z}$, these models usually use a subset of the full search space to approximate the posterior distribution, which means
\begin{equation}
   P(\mathbf{y} \vert \mathbf{x}) \simeq \sum_{\mathbf{z}\in B(\mathbf{z})} P(\mathbf{y} \vert \mathbf{z},\mathbf{x})P(\mathbf{z} \vert \mathbf{x}) 
\end{equation}
where $B(\mathbf{z})$ is a subset of the full search space.

\name also requires marginalizing over the latent states to compute the most likely translations. However, because the latent states don't directly depend on each other, inference in \name allows for exhaustive search over the latent states.
}

Finding the most likely translation using \name also requires marginalizing over the latent states. However, because the latent states in \name don't directly depend on each other, we can exhaustively search over the latent states. 
In particular, we can show that when $\mathbf{y}$ is fixed (observed),  the $\{\mathbf{z}_n\}_{n=1}^{N}$ variables are d-separated~\cite{bishop2006} i.e., are mutually independent. As a result, the time complexity for searching latent sequence $\mathbf{z}$ drops from ${|V_{\mathbf{z}}|}^N$ to $N|V_{\mathbf{z}}|$.

The posterior distribution for the translation probability at a time step $n$ can be computed as follows:
\begin{equation}
\begin{aligned}
    P&(\mathbf{y}_n \vert \mathbf{x}, \mathbf{y}_{<n}) \\
    &= \sum_{\mathbf{z}_n\in F(\mathbf{z}_n)} P(\mathbf{y}_n, \mathbf{z}_n \vert \mathbf{x}, \mathbf{y}_{<n})\\
    &= \sum_{\mathbf{z}_n\in F(\mathbf{z}_n)} P(\mathbf{y}_n \vert \mathbf{z}_n,\mathbf{x}, \mathbf{y}_{<n}) \times P(\mathbf{z}_n \vert \mathbf{x}, \mathbf{y}_{<n})
\end{aligned}
\end{equation}
where, $F(\mathbf{z}_n)$ is the full search space of latent states $\mathbf{z}_n$ and the joint probability is factorized as specified in Equation~\ref{eq:joint}.

For decoding words, we use standard beam search\footnote{Note our primary goal is to perform exhaustive search in the latent space. Search in the target vocabulary space remains exponential in our model.} with $P(\mathbf{y}_n, \mathbf{z}_{n} \vert \mathbf{x}, \mathbf{y}_{<n})$ as the beam cost. With this inference scheme, we can easily control decoding for diversity, by feeding different $\mathbf{z}$ sequences to the right branch of the decoder and decode diverse $\mathbf{y}_n$ by directly using $P(\mathbf{y}_n \vert \mathbf{z}_{n}, \mathbf{x}, \mathbf{y}_{<n})$ as the beam cost. Unlike the two-step decoding models which only evaluate a small number of $\mathbf{z}_n$ values at each time step (constrained by beam size), \name evaluates all possible values for $\mathbf{z}_n$ at each time step, while avoiding the evaluation of all possible sequences.

\subsection{Training with Neural Expectation Maximization}
The log-likelihood of \name's parameters $\boldsymbol{\theta}$\footnote{This includes the trainable parameters of the encoder, decoder, and the latent state embeddings.} computed on one training pair $\langle \mathbf{x}, \mathbf{y} \rangle \in D$ is given by:
\begin{equation}
\label{eq:loglikelihood}
\begin{aligned}
\mathcal{L}(\mathbf{\boldsymbol{\theta}}) &= \log P(\mathbf{y} \vert \mathbf{x}; \mathbf{\boldsymbol{\theta}}) \\
&= \sum_{n=1}^{N}\log P(\mathbf{y}_n \vert \mathbf{x}, \mathbf{y}_{<n}; \mathbf{\boldsymbol{\theta}}) \\
&= \sum_{n=1}^{N}\log \sum_{\mathbf{z}_n \in V_{\mathbf{z}}} P(\mathbf{y}_n, \mathbf{z}_n\vert \mathbf{x}, \mathbf{y}_{<n})
\end{aligned}
\end{equation}

Directly optimizing the log-likelihood function (equation~\ref{eq:loglikelihood}) with respect to model parameter $\boldsymbol{\theta}$ is challenging because of the highly non-convex function $P(\mathbf{y}_n, \mathbf{z}_n\vert \mathbf{x}, \mathbf{y}_{<n})$ and the marginalization over $\mathbf{z}_n$.\footnote{Note that marginalization is an issue during training, unlike in inference. As $P(y_n, z_n)$ is already an non-convex function with respect to $\theta$, summing $P(y_n, z_n)$ over different values of $z_n$ makes the function more complicated. Besides, we also need to compute gradients to update the parameters and computing the gradient of a log-of-sum function is costly and unstable. During the translation, we only need to compute the value of $\mathcal{L}(\theta)$ as score for beam searching. Therefore, the marginalization is not an issue.} Alternatively, we optimize the system parameters by Expectation Maximization (EM).

Using Jensen's inequality, equation~\eqref{eq:loglikelihood} can be re-written as:
\begin{equation}
\label{eq:lower}
\begin{aligned}
\mathcal{L}(\mathbf{\boldsymbol{\theta}}) &= \sum_{n=1}^N\log \sum_{\mathbf{z}_n \in V_{\mathbf{z}}} Q(\mathbf{z}_n) \frac{P(\mathbf{y}_n, \mathbf{z}_n\vert \mathbf{x}, \mathbf{y}_{<n})}{Q(\mathbf{z}_n)} \\
\geq & \sum_{n=1}^N \sum_{\mathbf{z}_n \in V_{\mathbf{z}}} Q(\mathbf{z}_n) \log \frac{P(\mathbf{y}_n, \mathbf{z}_n\vert \mathbf{x}, \mathbf{y}_{<n})}{Q(\mathbf{z}_n)} \\
=& \mathcal{L}_{lower}(Q,\boldsymbol{\theta})
\end{aligned}
\end{equation}
where $\mathcal{L}_{lower}(Q,\boldsymbol{\theta})$ is the lower bound of the log-likelihood and $Q$ is any auxiliary probability distribution defined on $\mathbf{z}_n$. $\boldsymbol{\theta}$ is omitted from the expression for simplicity.

We set $Q(\mathbf{z}_n)=P(\mathbf{z}_n \vert \mathbf{x}, \mathbf{y}_{\leq n}; \boldsymbol{\theta}^{old})$, the probability of the latent state computed by the decoder (shown in the left branch in Figure~\ref{fig:architecture}). Substituting this in equation~\eqref{eq:lower}, we obtain the lower bound as

\begin{equation}
\begin{aligned}
&\mathcal{L}_{lower}(Q,\boldsymbol{\theta})=\\ &\sum_{n=1}^N \sum_{\mathbf{z}_n \in V_{\mathbf{z}}} P(\mathbf{z}_n \vert \mathbf{x}, \mathbf{y}_{\leq n}; \boldsymbol{\theta}^{old})
\times \log P(\mathbf{y}_n, \mathbf{z}_n\vert \mathbf{x}, \mathbf{y}_{<n}; \boldsymbol{\theta}) \\
&- P(\mathbf{z}_n \vert \mathbf{x}, \mathbf{y}_{\leq n}; \boldsymbol{\theta}^{old}) \times \log P(\mathbf{z}_n \vert \mathbf{x}, \mathbf{y}_{\leq n}; \boldsymbol{\theta}^{old}) \\
&= \mathcal{Q}(\boldsymbol{\theta} , \boldsymbol{\theta}^{old}) + C
\end{aligned}
\end{equation}
where 

\begin{equation}
\begin{aligned}
\mathcal{Q}(\boldsymbol{\theta}, \boldsymbol{\theta}^{old}) =& \sum_{n=1}^N \sum_{\mathbf{z}_n \in V_{\mathbf{z}}} P(\mathbf{z}_n \vert \mathbf{x}, \mathbf{y}_{\leq n}; \boldsymbol{\theta}^{old}) \\
& \times \log P(\mathbf{y}_n, \mathbf{z}_n\vert \mathbf{x}, \mathbf{y}_{<n}; \boldsymbol{\theta})
\end{aligned}
\label{eq:expectation}
\end{equation}

EM algorithm for optimizing $\mathcal{Q}(\boldsymbol{\theta}, \boldsymbol{\theta}^{old})$ consists of two major steps.
In the E-step, we compute the posterior distribution of $\mathbf{z}_n$ with respect to $\boldsymbol{\theta}^{old}$ by

\begin{equation}
\begin{aligned}
\gamma&(\mathbf{z}_n = i) = P(\mathbf{z}_n = i \vert \mathbf{x}, \mathbf{y}_{\leq n}) \\
&= \frac{P(\mathbf{y}_n, \mathbf{z}_n = i\vert \mathbf{x}, \mathbf{y}_{<n})}{\sum_{\mathbf{z}_n = j}P(\mathbf{y}_{n}, \mathbf{z}_n = j\vert \mathbf{x}, \mathbf{y}_{<n})}
\end{aligned}
\end{equation}

where $\gamma(\mathbf{z}_n = i)$ is the responsibility of $\mathbf{z}_n = i$ given $\mathbf{y}_{n}$, and can be calculated by equation~\eqref{eq:joint}.

In the M-step, we aim to find the configuration of $\boldsymbol{\theta}$ that would maximize the expected log-likelihood using the posteriors computed in the E-step. In conventional EM algorithm for shallow probabilistic graphical model, the M-step is generally supposed to have closed-form solution. However, we model the probabilistic dependencies by deep neural networks, where $\mathcal{Q}(\boldsymbol{\theta}, \boldsymbol{\theta}^{old})$ is highly non-convex and non-linear with respect to network parameters $\boldsymbol{\theta}$. Therefore, there exists no analytical solution to maximize it. However, since deep neural network is differentiable, we can update $\boldsymbol{\theta}$ by taking a gradient ascent step:

\begin{equation}
\begin{aligned}
	\boldsymbol{\theta}^{new}=\boldsymbol{\theta}^{old} + \eta \frac{\partial \mathcal{Q}( \boldsymbol{\theta},\boldsymbol{\theta}^{old})}{\partial \boldsymbol{\theta}},
\end{aligned}
\label{eq:theta}
\end{equation}

The resulting algorithm belongs to the class of \textit{generalized EM algorithms} and is guaranteed (for a sufficiently small learning rate $\eta$) to converge to a (local) optimum of the data log likelihood \cite{wu83}.

\subsection{Regularized EM training}
The EM training we derived does not assume any supervision for the latent variables $\mathbf{z}$. This can be seen as inferring the latent syntax of the target sentences by clustering the target side tokens into $|V_{\mathbf{z}}|$ different categories. Given some token-level syntactic information, we can modify the training procedure to regularize the generation of latent sequence $P(\mathbf{z}_n \vert \mathbf{x}, \mathbf{y}_{< n})$ such that true latent sequences have higher probabilities under the model. In this work, we consider parts-of-speech sequences of the target sentences for regularization. 

The regularized EM training objective is thus redefined as
\begin{align}
	\mathcal{L}_{total}(\mathbf{\boldsymbol{\theta}}) = \mathcal{L}_{lower}(\boldsymbol{\theta}) + \lambda \mathcal{L}_{\mathbf{z}}(\boldsymbol{\theta}),
\end{align} 
where $\mathcal{L}_{lower}(\boldsymbol{\theta})$ is the EM lower bound in equation~\eqref{eq:lower} and $\mathcal{L}_{\mathbf{z}}(\boldsymbol{\theta})$ denotes cross entropy loss between $P(\mathbf{z}_n \vert \mathbf{x}, \mathbf{y}_{< n})$ and the true POS tag sequences and $\lambda$ is a hyper-parameter that controls the impact of the regularization.

This regularized training algorithm is shown in Algorithm~\ref{algo:training}.

\begin{algorithm}
\small
\caption{Training NMT with latent POS tag sequences through Regularized Neural EM}
\begin{algorithmic}
\STATE \textbf{Objective:} Maximize the log likelihood function $\mathcal{Q}( \boldsymbol{\theta},\boldsymbol{\theta}^{old})$ with respect to $\mathbf{\boldsymbol{\theta}}$ over observed variable $\mathbf{y}$ and latent variable $\mathbf{z}$, governed by parameters $\boldsymbol{\theta}$.
\STATE \textbf{Input:} Parallel corpus training data $\langle \mathbf{x}, \mathbf{y} \rangle$; POS tag sequence $\mathbf{z}$ of target sentence; the number of EM update steps per batch $K$.
\STATE \textbf{Initialize:} Initialize random values for the parameters $\boldsymbol{\theta}^{old}$.
\WHILE{Training loss has not converged}
	\STATE Select $\langle \mathbf{x}, \mathbf{y} \rangle \in D$, parse $\mathbf{z}$ of $\mathbf{y}$.
	\FOR{$k \gets 1$ to $K$}
		\STATE \textbf{1. E-step:} Evaluate $\gamma(\mathbf{z}_n=i)$ for $i\leq |V_{\mathbf{z}}|$.
		\STATE \textbf{2. M-step:} Evaluate $\boldsymbol{\theta}^{new}$ given by equation~(\ref{eq:theta}).
		\STATE \textbf{3.} Let $\boldsymbol{\theta}^{old} \gets \boldsymbol{\theta}^{new}$.
	\ENDFOR
\ENDWHILE
\end{algorithmic}
\label{algo:training}
\end{algorithm}

\section{Evaluation}
We evaluate \name on four translation tasks, including three with moderate sized datasets IWSLT 2014~\cite{Cettolo14} German-English (De-En), English-German (En-De), English-French (En-Fr), and one with a relatively larger dataset, the WMT 2014 English-German (En-De). We describe the datasets in more details in the appendix.

We compare against three types of baselines: (i) general Seq2Seq models that use no syntactic information, (ii) models that incorporate source side syntax directly, and multitask learning models which include syntax indirectly, and (iii) models that use syntax on the target side. We also define a \name \textbf{Empirical Upper Bound (EUB)}, which is our proposed model using true POS tag sequences for inference.

We use BLEU as the evaluation metric~\cite{papineni02} for translation quality. For diverse translation evaluation, we utilize \textit{distinct-1} score~\cite{li2016diversity} as the evaluation metric, which is the number of distinct unigrams divided by total number of generated words.

For all translation tasks, we choose the \textit{base} configuration of Transformer with $d_{model}=512$.
During training, we choose Adam optimizer~\cite{Kingma14} with $\beta_1 = 0.9$, $\beta_2 = 0.98$ with initial learning rate is 0.0002 with 4000 warm-up steps. We describe additional implementation and training details in the Appendix.
\eat{\subsection{Implementation and Training Details}
We implement all Transformer-based models using \texttt{Fairseq}~\footnote{https://github.com/pytorch/fairseq} Pytorch framework. 

For all translation tasks, we choose the \textit{base} configuration of Transformer with $d_{model}=512$.
During training, we choose Adam optimizer~\cite{Kingma14} with $\beta_1 = 0.9$, $\beta_2 = 0.98$. The initial learning rate is 0.0001 with 4000 warm-up steps. The learning rate is scheduled with the same rule as in~\cite{Vaswani17}. Each batch on one GPU contains roughly 1700 tokens for IWSLT tasks and 500 tokens for the WMT En-De task. We train IWSLT tasks using two 1080Ti GPUs and train WMT task using 8 K80 GPUs. The hyperparameter $\lambda$ is set to 0.5. For inference, we use beam search with beam size 5 to generate candidates. 
}

\subsection{Results on IWSLT'14 Tasks}

Table~\ref{tab:result-iwslt} compares \name versions against some of the state-of-the-art models on the IWSLT'14 dataset.
\name-K rows show results when varying the number of EM update steps per batch ($K$).

On the De-En task, \name provides a 1.7 points improvements over the Transformer baseline, demonstrating that the \name's improvements come from incorporating target side syntax effectively. This result is also better than a transformer-based source side syntax model by 1.5 points.
\name results are also better than the published numbers for LSTM-based models that use multi-task learning for source side and models that uses target side syntax. Note that since the underlying architectures are different, we only present these results to show that the results with \name are comparable with other models that have incorporated syntax.

On the En-De task, our model achieves 29.2 in terms of BLEU score, with 2.6 points improvement over the Transformer baseline and 2.4 points improvement over Transformer-based Source Side Syntax model. Compared with NPMT~\cite{huang2018towards}, which is a BiLSTM based model, we achieve 3.84 point improvement.

On the En-Fr task, our model set a new state-of-the-art with a BLEU score of 40.6, which is 1.7 points improvement over the second best model which uses Transformer to incorporate source side syntax knowledge. Our model also surpasses the basic Transformer model by about 2.1 points.

We notice that across all tasks, the performance of our model improves with number of EM update steps per batch ($K$). With larger $K$ values, we get better lower bounds $\mathcal{L}_{lower}(\boldsymbol{\theta})$ on each training batch, thus leading to better optimization. For update steps beyond $K>5$, the performance does not improve any more.

Last, the EUB row indicates the performance that can be obtained when feeding in the true POS tags. The large improvement here shows the potential for improvement when modeling target side syntax.

\begin{table*}[!t]
\centering
\small
\begin{tabular}{cccccl}
\toprule
\multirow{2}{*}{\textbf{Method Type}} & \multirow{2}{*}{\textbf{Model}} & \multicolumn{3}{c}{\textbf{BLEU}} \\\cline{3-5}
& & \textbf{De-En} & \textbf{En-De} & \textbf{En-Fr}     \\
\midrule
\multirow{4}{*}{BiLSTM} & BiLSTM~\cite{denkowski2017} & -- & -- & 34.8 \\
& Dual Learning~\cite{Wang18dual} & 32.35 & -- & -- \\
& AST~\cite{Cheng18Towards} & -- & -- & 38.03 \\
& NPMT~\cite{huang2018towards} & -- & 25.36 & -- \\
\midrule
Multi-Task (BiLSTM) & MTL-NMT~\cite{Niehues17} & 27.78 & -- & -- \\
\midrule
Source Side Syn. (Transformer) & Source-NMT~\cite{SennrichLinguistic16} & 33.5 & 26.8 & 38.9 \\
\midrule
\multirow{2}{*}{Target Side Syn. (BiLSTM)} & DSP-NMT~\cite{Shu18} & 29.78 & -- & -- \\
& Tree-decoder~\cite{wang2018tree} & 32.65 & -- & -- \\
\midrule
\multirow{6}{*}{Transformer}& Transformer  & 33.3 & 26.6 & 38.5 \\
& \name (Unsupervised)  & 30.8 & 25.2 & 34.5 \\
& \name (K=1)  & 34.63 & 28.1 & 39.7 \\
& \name (K=3)  & 34.91 & 28.9 & 40.4\\
& \name (K=5)  & \textbf{35.0} & \textbf{29.2} & \textbf{40.6} \\
& \name \textbf{EUB} & 51.4 & 47.3 & 54.2\\
\bottomrule
\end{tabular}
\caption{\textbf{IWSLT'14 English-German and English-French results} - shown are the BLEU scores of various models on TED talks translation tasks. We highlight the \textbf{best} model in bold.}
\label{tab:result-iwslt}
\end{table*}

Table~\ref{tab:result-de-en-samples} shows one example where \name produces correct translations for a long input sentence. The output of \name is close to the reference and the output of \name when given the gold POS tag sequence is even better, demonstrating the benefits of modeling syntax. The transformer model however fails to decode the later portions of the long input accurately. 
\begin{table*}[!t]
\centering
\small
\begin{tabular}{cp{13cm}cl}
\toprule
SRC & letztes jahr habe ich diese beiden folien gezeigt , um zu veranschaulichen , dass die arktische eiskappe , die für annähernd drei millionen jahre die grösse der unteren 48 staaten hatte , um 40 prozent geschrumpft ist . \\
REF & last year i showed these two slides so that demonstrate that the arctic ice cap , which for most of the last three million years has been the size of the lower 48 states , has shrunk by 40 percent . \\
\midrule
Transformer & last year , i showed these two slides to illustrate that the arctic ice caps that had the size of the lower 48 million states \color{red}{to 40 percent .} \\
\name  & last year , i showed these two slides to illustrate that the arctic ice cap , \color{blue}{which for nearly three million years} had the size of the lower 48 states , \color{blue}{was shrunk by 40 percent .}  \\
\name (groundtrue POS)  & last year i showed these two slides just to illustrate that the arctic ice cap , \color{blue}{which for nearly about the last three million years} has been the size of the lower 48 states , \color{blue}{has shrunk by 40 percent .} \\
\bottomrule
\end{tabular}

\caption{Translation examples on IWSLT'14 De-En dataset from our model and the Transformer baseline. We put correct translation segment in blue and highlight the wrong one in red.}
\label{tab:result-de-en-samples}
\end{table*}

\subsection{Speed}
We compare the speeds of our (un-optimized) implementation of \name with a vanilla transformer with no latents in its decoder. Table~\ref{tab:result-speed} shows the training time per epoch, and the inference time for the whole test set. computed on the IWSLT'14 De-En task. When $K=1$, \name takes almost twice as much time as the vanilla transformer for training. Increasing $K$ increases training time further. For inference, \name takes close to four times as much time compared to the vanilla Transformer. In terms of complexity, \name only adds a linear term (in POS tag size to the decoding complexity. Specifically, its decoding complexity is $B \times O(m) \times O(\vert V_{\mathbf{z}}\vert \times N)$ where $B$ is beam size, $m$ is a constant proportional to the tag set size and $N$ is output size. 
As the table shows, empirically, our current implementation incurs $m \simeq 4$. We leave further optimizations for future work.
\eat{We find that our current implementation of the \texttt{Add} operation is the main bottleneck due to its high memory cost, which we believe can be optimized to further improve efficiency.
\rev{Xuewen, do you know if this can be reduced or are you guessing.}}

\begin{table}[!t]
\centering
\small
\begin{tabular}{cccl}
\toprule
\textbf{Model} & \textbf{Training Time/Epoch}$\downarrow$ & \textbf{Inference Time}$\downarrow$ \\
\midrule
Transformer & \textbf{3.6} min & \textbf{12.8} s\\
\name (K=1) & 6.3 min & 56.1 s \\
\name (K=3) & 18.1 min & 55.6 s \\
\name (K=5) & 28.0 min & 55.6 s \\
\bottomrule
\end{tabular}
\caption{\textbf{IWSLT'14 De-En training and inference speed evaluation}. $\downarrow$ means the smaller the better. We highlight the \textbf{best} model in bold.}
\label{tab:result-speed}
\end{table}

\subsection{Diversity}

We compare the diversity of translations using \textit{distinct-1} score~\cite{li2016diversity}, which is simply the number of distinct unigrams divided by total number of generated words. We use our model to generate 10 translations for each source sentence of the test dataset. We then compare our results with baseline Transformer. The result is shown in Table~\ref{tab:result-diversity}. Much like translation quality, \name's diversity increases with number of EM updates and is better than diversity of the transformer and a source side encoder model.

\begin{table}[!t]
\centering
\small
\begin{tabular}{ccccl}
\toprule
\multirow{2}{*}{\textbf{Model}} & \multicolumn{3}{c}{\textbf{distinct-1}} \\\cline{2-4}
& De-En & En-De & En-Fr \\
\midrule
Transformer & 0.231 & 0.242 & 0.258\\
Source-NMT & 0.232 & 0.244 & 0.260 \\
\name (Unsupervised) & 0.228 & 0.231 & 0.239 \\
\name (K=1) & 0.237 & 0.251 & 0.265 \\
\name (K=3) & 0.241 & 0.253 & 0.270 \\
\name (K=5) & \textbf{0.245} & \textbf{0.255} & \textbf{0.273} \\
\name \textbf{EUB} & 0.328 & 0.516 & 0.354 \\
\bottomrule
\end{tabular}
\caption{\textbf{IWSLT'14 En-De/De-En/En-Fr diversity translation evaluation}. We highlight the \textbf{best} model in bold.}
\label{tab:result-diversity}
\end{table}

\subsubsection{Controlling Diversity with POS Sequences}
One of the main strengths of \name is that it can generate translations conditioned on a given POS sequence. First, we present some examples that we generate by decoding over different POS tag sequences. Given a source sentence, we use \name to provide the most-likely target pos tag sequence. Then, we obtain a random set of valid POS tag sequences that differ from this maximum likely sequence by some edit distance. For each of these randomly sampled POS tag sequences, we let \name generate a translation that fits the sequence. Table~\ref{tab:result-de-en-diverse} shows some example sentences. \name is able to generate diverse translations that reflect the sentence structure implied by the input POS tags. 
However, in trying to fit the translation into the specified sequence, it deviates somewhat from the ideal translation. 

\begin{figure}[t!]
\centering
\includegraphics[width=0.45\textwidth]{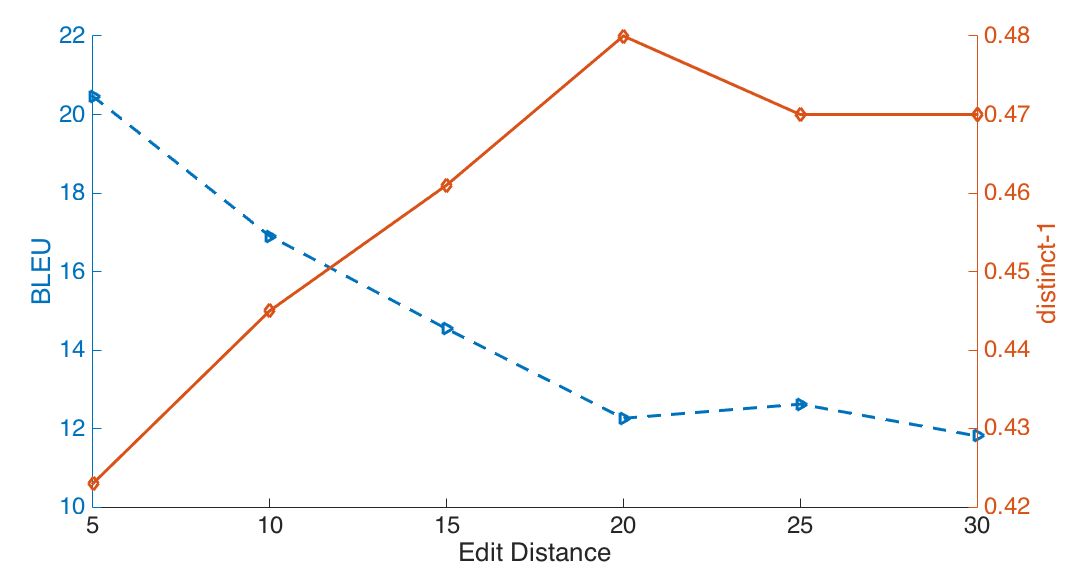}
\caption{Diversity vs. Translation Quality: BLEU and \textit{distinct-1} scores for targets decoded using POS sequences of increasing edit distance.}
\label{fig:edit}
\vspace{-1em}
\end{figure}

To understand how diversity plays against translation quality, we also conduct a small scale quantitative evaluation. We pick a subset of the test dataset, and for each source sentence in this subset, we sample $10$ POS tag sequences whose edit distance to their corresponding Top-1 POS tag sequence equal to a specific value, we then use them to decode $W$ translations. We calculate their final BLEU and \textit{distinct-1} scores. 
The results are shown in Figure:~\ref{fig:edit}. As the edit distance increases, diversity increases dramatically but at the cost of translation quality. Since the POS tag sequence acts as a template for generation, as we move further from the most likely template, the model struggles to fit the content accurately. Understanding this trade-off can be useful for re-ranking or other scoring functions. 

\begin{table}[!t]
\centering
\small
\begin{tabular}{cp{6.7cm}cl}
\toprule
SRC & und natürlich auch , wie nimmt gestaltung einfluss auf die wahrnehmung . \\
REF & and of course how design affects perception . \\
\midrule
$0$ & cc in nn wrb nn vbz vbz nn . \\
 &  and of course how design is affecting perception . \\
$20$ &  ls rb vb in dt nn vbz jj jj jj . \\
 & i also think that the design is affecting perception . \\
 $30$ & rb , dt nn nn dt nn in prp rb . \\
 & also , the way design adds influence in perception too . \\
 $30$ & prp vbz in prp vb vbn rb in nn . \\
 & it turns out it included design impact on perception .\\
\bottomrule
\end{tabular}
\caption{Examples of translations decoded from specified POS sequences with different edit distances (shown as values in first column). SRC: source sentence. REF: reference sentence.}
\label{tab:result-de-en-diverse}
\end{table}

\subsection{Results on WMT'14 En-De}
To assess the impact on a larger dataset, we show results on the WMT'14 English-German in table~\ref{tab:result-wmt-en-de}. Compared to the previously reported systems, we see that our transformer implementation is a strong baseline.
\name produces small gains, with the best gain at K=5 -- a BLEU score improvement of 0.6. This demonstrates that syntactic information can contribute more to the increase of translation quality on a smaller dataset. 


\eat{While large data settings often favor models with less inductive bias, in this particular case we need further experiments with higher K settings to draw clear conclusions.}

\begin{table}[!t]
\centering
\begin{tabular}{ccl}
\toprule
\textbf{Model} & \textbf{BLEU} \\
\midrule
BiRNN+GCN~\cite{bastings2017graph} & 23.9 \\
ConvS2S~\cite{Gehring17} & 25.16  \\
MoE~\cite{Shazeer17} & 26.03 \\
Transformer (base) & 27.3 \\
\name (K=1) (base) & 27.6 \\
\name (K=3) (base) & 27.8 \\
\name (K=5) (base) & \textbf{27.9} \\
\bottomrule
\end{tabular}

\caption{\textbf{WMT'14 English-German results} - shown are the BLEU scores of various models on TED talks translation tasks. We highlight the \textbf{best} model in bold.}
\label{tab:result-wmt-en-de}
\end{table}

\section{Related Work}
Attention-based Neural Machine Translation (NMT) models have shown promising results in various large scale translation tasks~\cite{BahdanauCB14, luong2015, SennrichHB16, Vaswani17} using an \texttt{Seq2Seq} structure. Many Statistical Machine Translation (SMT) approaches have leveraged benefits from modeling syntactic information~\cite{Chiang2009, Huang2006, Shen2008}. Recent efforts have demonstrated that incorporating syntax can also be useful in neural methods as well.

One branch uses features on the source side to help improve the translation performance~\cite{SennrichLinguistic16, Morishita18, eriguchi2016tree}. Sennrich \textit{et al.}~\shortcite{SennrichLinguistic16} explored linguistic features like lemmas, morphological features, POS tags and dependency labels and concatenate their embeddings with sentence features to improve the translation quality. In a similar vein, Morishita \textit{et al.}~\shortcite{Morishita18} and Eriguchi \textit{et al.}~\shortcite{eriguchi2016tree}, incorporated hierarchical subword features and phrase structure into the source side representations. Despite the promising improvements, these approaches are limited in that the trained translation model requires the availability of external tools during inference -- the source text needs to be processed first to extract syntactic structure~\cite{eriguchi2017learning}.

Another branch uses multitask learning, where the encoder of the NMT model is trained to produce multiple tasks such as POS tagging, named-entity recognition, syntactic parsing or semantic parsing~\cite{eriguchi2017learning, Baniata18, Niehues17, Zaremoodi18}. These can be seen as models that implicitly generate syntax informed representations during encoding. With careful model selection, these methods have demonstrate some benefits in NMT.

The third branch directly models the syntax of the target sentence during decoding~\cite{gu2018top, wang2018tree, wu2017sequence, aharoni2017, bastings2017graph, li2018target}. 
Aharoni \textit{et al.}~\shortcite{aharoni2017} treated constituency trees as sequential strings and trained a \texttt{Seq2Seq} model to translate source sentences into these tree sequences. Wang \textit{et al.}~\shortcite{wang2018tree} and Wu \textit{et al.}~\shortcite{wu2017sequence} proposed to use two RNNs, a Rule RNN and a Word RNN, to generate a target sentence and its corresponding tree structure. Gu \textit{et al.}~\shortcite{gu2018top} proposed a model to translate and parse at the same time. 
However, apart from the complex tree structure to model, they all have a similar architecture as shown in Figure~\ref{fig:Bayes2}, which limits them to only exploring a small portion of the syntactic space during inference.

\name uses simpler parts-of-speech information in a latent syntax model, avoiding the typical exponential search complexity in the latent space with a linear search complexity and is optimized by EM. This allows for better translation quality as well as diversity. Similar to our work, \cite{Shankar2018Surprisingly} and \cite{shankar2018posterior} proposed a latent attention mechanism to further reduce the complexity of model implementation by taking a top-K approximation instead of the EM algorithm as in \name.

\section{Conclusion}
Modeling target-side syntax through true latent variables is difficult because of the additional inference complexity. In this work, we presented \name, a latent syntax model that allows for efficient exploration of a large space of latent sequences. This yields significant gains on four translation tasks, IWSLT'14 English-German, German-English, English-French and WMT'14 English-German. The model also allows for better decoding of diverse translation candidates. This work only explored parts-of-speech sequences for syntax. Further extensions are needed to tackle tree-structured syntax information. 


\section{Acknowledgements}
This work is supported in part by the National Science Foundation under Grants NSF CNS 1526843 and IIS 1815358. We thank Jiangbo Yuan, Wanying Ding, Yang Liu, Guillaume Rabusseau and Jeffrey Heinz for valuable discussions.
\bibliography{emnlp-ijcnlp-2019}
\bibliographystyle{acl_natbib}
\end{document}


\maketitle

\section{Appendix}

\subsection{Implementation and Training Details}
We implement all Transformer-based models using \texttt{Fairseq}~\footnote{https://github.com/pytorch/fairseq} Pytorch framework. 

For all translation tasks, we choose the \textit{base} configuration of Transformer with $d_{model}=512$.
During training, we choose Adam optimizer~\cite{Kingma14} with $\beta_1 = 0.9$, $\beta_2 = 0.98$. The initial learning rate is 0.0002 with 4000 warm-up steps. The learning rate is scheduled with the same rule as in~\cite{Vaswani17}. Each batch on one GPU contains roughly 2000 tokens for IWSLT tasks and 800 tokens for the WMT En-De task. We train IWSLT tasks using two 1080Ti GPUs and train WMT task using 8 K80 GPUs. The hyperparameter $\lambda$ is set to 0.2. For inference, we use beam search with beam size 5 to generate candidates. 
\subsection{Dataset Details}
We evaluate our model on two small translation datasets - IWSLT'14 German-English (De-En) and English-French (En-Fr)~\cite{Cettolo14} and a much bigger one - WMT'14 English-German (En-De). 

\textbf{IWSLT'14 En-De/En-Fr} 
We use the datasets extracted from IWSLT 2014 machine translation evaluation campaign~\cite{Cettolo14}, which consists of 153K/220K training sentence pairs for En-De/En-Fr tasks. For En-De, we use 7K data split from the training set as the validation set and use the concatenation of dev2010, tst2010, tst2011 and tst2012 as the test set, which is widely used in prior studies~\cite{huang2018towards, Tianyu19, bahdanau2016actorcritic, Ranzato16}.
For En-Fr, the tst2014 is taken as the validation set and tst2015 is used as the test set, which is the same with prior studies~\cite{denkowski2017, Cheng18Towards}.
We also lowercase the sentences of En-De and En-Fr following general practice. Before encoding sentences using sub-word types based on byte-pair encoding~\cite{sennrich2016neural}, which is a common practice in NMT, we parse POS tag sequences of the sentences using Stanford Parser~\cite{Chen14}. The POS tag sequences produce POS vocabulary of size 32 for both English and French and 32 for German. Sentences are then encoded using sub-word types. To make the lengths of POS tag sequences equal to their corresponding sub-word sentences, if several sub-words belong to the same word, they are given the same POS tag. For IWSLT'14 En-De dataset, we build a English sub-word vocabulary of size 6632 and a German sub-word vocabulary of size 8848. For En-Fr dataset, we build a English sub-word vocabulary of size 7172 and a French sub-word vocabulary of size 8740.

\textbf{WMT'14 English-German (En-De)} 

We use the same dataset as~\cite{Vaswani17}, which consists of 4.5M sentence pairs. We use the concatenation of newstest2012 and newstest2013 as the validation set and newstest2014 as the test set. Sentences are encoded using byte-pair encoding with a shared vocabulary of about 40K sub-word tokens. The method to generate POS tag sequences is the same, except that we merge some POS tags of similar meaning to one and get a POS tag vocabulary of size 16 for both German and English. This operation reduces computational cost, and gives us a bigger batch for training.
\bibliography{emnlp-ijcnlp-2019}
\bibliographystyle{acl_natbib}